\documentclass[letterpaper, 10 pt, conference]{ieeeconf}  

\IEEEoverridecommandlockouts                              
\usepackage{times}  
\usepackage{helvet}  
\usepackage{courier}  
\usepackage[normalem]{ulem}
\usepackage{url}  
\usepackage{graphicx}  
\usepackage{amsmath}
\usepackage{booktabs}
\usepackage{subcaption}
\usepackage{placeins}
\usepackage{balance} 
\usepackage{color}
\usepackage{float}

\title{\LARGE \bf
DeepMoTIon: Learning to Navigate Like Humans
}

\author{Mahmoud Hamandi$^{1}$, Mike D'Arcy$^{2}$, and Pooyan Fazli$^{3}$
\thanks{$^{1}$Mahmoud Hamandi is with LAAS-CNRS, Universit\'e de Toulouse, CNRS, Toulouse, France
        {\tt\small mhamandi@laas.fr}}%
\thanks{$^{2}$Mike D'Arcy is with the Department of Computer Science, Northwestern University, Evanston, IL, USA
        {\tt\small m.m.darcy@u.northwestern.edu}}%
\thanks{$^{3}$Pooyan Fazli is with the Department of Computer Science, San Francisco State University,
        San Francisco, CA, USA
        {\tt\small pooyan@sfsu.edu}}%
}

\begin{document}

\maketitle
\thispagestyle{empty}
\pagestyle{empty}

\begin{abstract}

We present a novel human-aware navigation approach, where the robot learns to mimic humans to navigate safely in crowds. The presented model, referred to as DeepMoTIon, is trained with pedestrian surveillance data to predict human velocity in the environment. The robot processes LiDAR scans via the trained network to navigate to the target location. We conduct extensive experiments to assess the components of our network and prove their necessity to imitate humans. Our experiments show that DeepMoTIion outperforms all the benchmarks in terms of human imitation, achieving a 24\% reduction in time series-based path deviation over the next best approach. In addition, while many other approaches often failed to reach the target, our method reached the target in 100\% of the test cases while complying with social norms and ensuring human safety.

\end{abstract}

\section{Introduction}
Robots are gradually moving from factories and labs to streets, homes, offices, and healthcare facilities. These robots are currently assigned tasks that require interaction with humans, such as guiding passengers through busy airport terminals \cite{triebel2016spencer} or roaming around university buildings and interacting with nearby humans \cite{FSS1716022,veloso2012cobots}.

As robots are increasingly becoming part of our everyday lives, it is essential for them to be aware of the surrounding humans while performing their tasks. Navigation is a basic skill for autonomous robots, but many traditional algorithms, such as A* and D*, do not consider the fact that the obstacles in the environment may be humans. While maneuvers made by these algorithms may produce short paths and avoid direct collisions, they do not consider social norms, such as walking on the right side and passing on the left. This can cause inconvenience for humans. We define human-aware navigation as the ability of the robot to navigate while complying with social norms and ensuring human safety.

While many existing systems allow robots to navigate safely within crowds \cite{ferrer2017robot,sisbot2007human}, they still rely heavily on manually crafted models of human motion. Such models may capture the aspects of human motion as understood by their designers, while they may likely miss subtle trends that characterize their human aspect. In addition, manually crafted models do not have a way to automatically adapt to different cultures, so it may require significant manual effort to be used in a different environment.

We present DeepMoTIon (Deep Model for Target-driven Imitation), a deep imitation learning algorithm that eliminates the need for an explicit model of human motion and instead learns the human navigation patterns directly by observing pedestrians. By imitating the motion patterns learned from real humans, the algorithm naturally follows social norms without needing such rules to be manually specified. Moreover, the network learns to decide on the direction and speed associated with raw LiDAR data without any preprocessing. The network is trained to learn the possible motion patterns it might face in human crowds on its own.

The goal of this work is to learn directly from pedestrian data without the need for a predefined human model. With the absence of a true model, learning the reward governing human motion is not feasible with current Inverse Reinforcement Learning algorithms such as the one presented in~\cite{ziebart2008maximum}. Our method tackles the imitation problem as a classification one, where the network learns a specific command for each observation without simulating the learned policy. This approach reduces the amount of time required for each architecture test and allows us to explore multiple network configurations.

The contributions of the paper are as follows: 
\begin{enumerate}
    \item We present a deep imitation learning algorithm to generate navigational commands and plan a path to the target in the environment, similar to humans, for a mobile ground robot. The proposed method outperforms all the benchmarks on time series-based path deviation and reaches the target in 100\% of the test cases while complying with social norms and ensuring human safety.
     \item Further, we  present  a  novel  loss function  to  train  the  network.  The  loss  function allows us to accommodate for human motion stochasticity while at the same time enabling the robot to navigate safely.
\end{enumerate}    

We conduct extensive experiments to assess the components of our deep neural network and prove their necessity to imitate humans.

\section{Background and State of the Art}
Previous work on human-aware navigation suggested to apply handcrafted models to control a robot about humans, define human-centric cost maps, or even follow humans through crowds. 
 
Helbing and Molar \cite{helbing1995social} presented the Social Force Model (SFM), where they modeled the assumed social forces governing the human motion. Ferrer \textit{et al.} \cite{ferrer2017robot} used the social force model to navigate in a way similar to humans. In their work, the robot navigates to the target while abiding by the social forces, that is, the robot is attracted by its target and repelled by pedestrians and obstacles. Furthermore, they extended the social force model to allow the robot to escort a human while providing a scheme to learn the parameters of the model.

Inverse reinforcement learning (IRL) has also been investigated to learn human-like navigation policies from data. Henry \textit{et al.} \cite{henry_learning_2010} adapted the MaxEnt IRL algorithm to partially-observed environments for socially-aware navigation. Vasquez \textit{et al.} \cite{vasquez2014inverse} tested a variety of features, such as crowd density and the social forces to learn a cost map that replicates the reward maximized in human navigation. Kim and Pineau \cite{kim2016socially} developed a navigation system based on maximum a posteriori Bayesian IRL. Using IRL for human-like navigation is a two-step process: first, training a reward function from human data using IRL, and then using a separate algorithm to find actions for the robot that are optimal under the learned reward function. On the other hand, in DeepMoTIon, the network is directly trained to produce actions that lead to human-like navigation.

Bera \textit{et al.} \cite{bera2017sociosense} predicted human motion after observing a set of related psychological cues, such as aggressiveness, tension, and level of activity. 
The robot then deduces the acceptable path from the predicted human locations and social distances inferred from the same psychological features.

Another approach to human-aware navigation was presented by Mehta \textit{et al.} \cite{mehta2016autonomous}, where the robot follows a human through crowds when it cannot navigate on its own. In their approach, the robot decides to navigate freely when the scene is clear or compromise its optimal shortest path by following a human to its goal. When neither possibility is viable, the robot stops and waits for a clearance.

Sisbot \textit{et al.} \cite{sisbot2007human} suggested a set of human-centric costs that allow the robot to navigate safely around humans. The method applied a cost-based navigation algorithm with a Gaussian coercing a safety distance about each human. In addition, the robot attempted to stay in the visual range of the existing pedestrians and to increase its own visibility near hidden areas, such as when rotating around a corner. 

Lu and Smart \cite{lu2013towards} proposed another method where the robot navigates following a human-aware cost map. Their approach forces the robot to navigate on the right side of a hallway, allowing opposing humans to navigate on its left. In addition, the robot communicated its awareness of the nearby pedestrians by tilting its head toward their eyes.

\begin{figure}[t]
      \centering
      \includegraphics[scale=0.20]{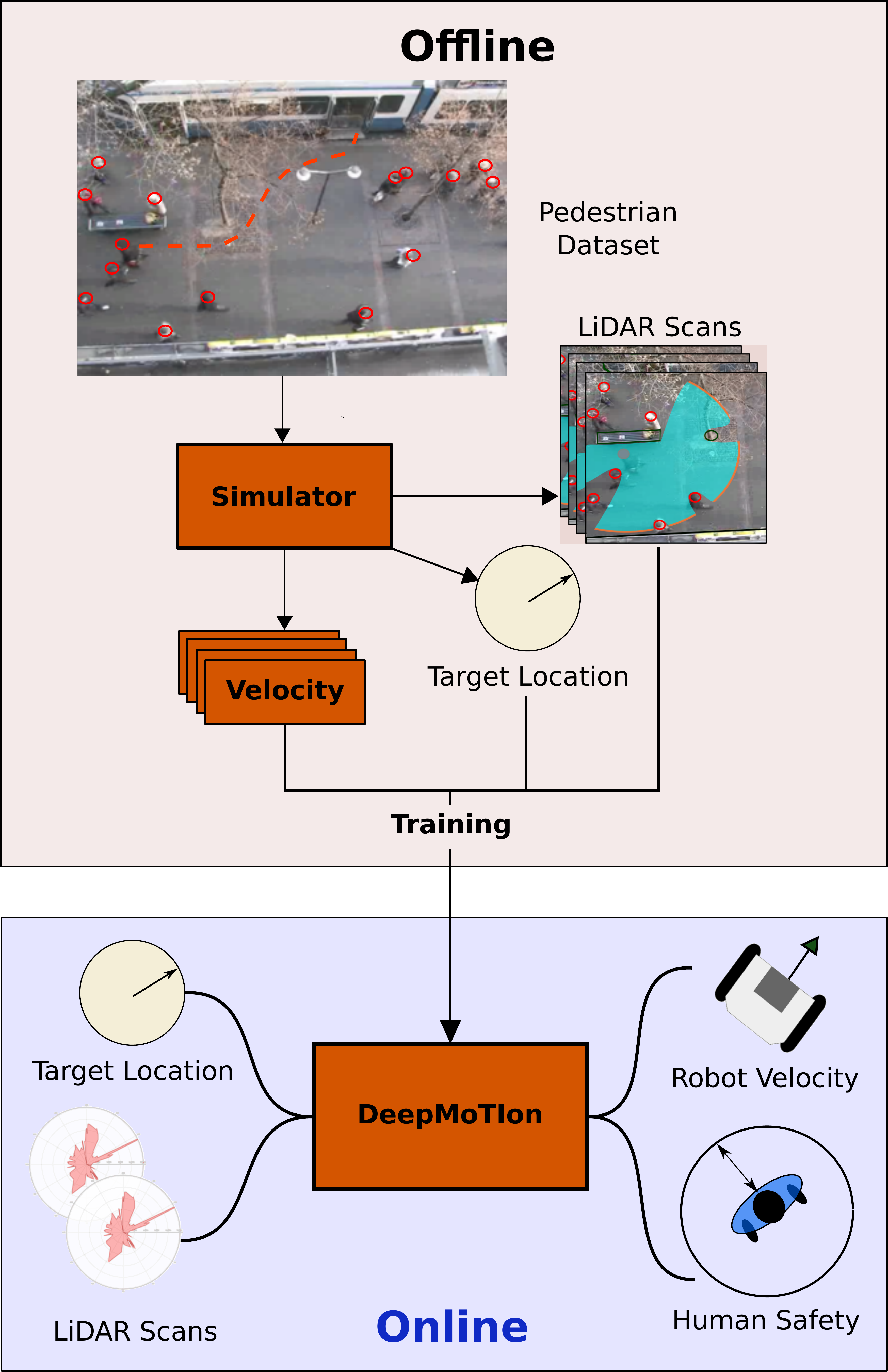}
      \caption{Algorithm overview: In the offline phase, we train a deep neural network, called DeepMoTIon, based on a pedestrian dataset. In the online phase, the network is given the target location and the last two consecutive LiDAR scans of the environment, and the network produces speed and direction outputs that are safe and adhere to social norms.}
      \label{ALG}
   \end{figure}

While these methods provide a model for human motion, multiple deep learning architectures were also presented in the literature to learn any navigation algorithm.
Pfeiffer \textit{et al.} \cite{pfeiffer2017perception} proposed an end-to-end network that allowed the robot to navigate based on LiDAR scans and target position. Similarly, Groshev \textit{et al.} \cite{groshev2017learning} presented a network that learns reactive policies that imitate a planning algorithm when provided with current and goal observations. Both papers presented novel ideas, however, they learn reactive policies ignoring previous robot states while trying to imitate long-term planning algorithms. 

Chen et al. \cite{chen_socially_2017} propose a socially-aware navigation approach using deep reinforcement learning. However, the socially-aware behavior of the method was achieved using a system of handcrafted rewards when training the model. This contrasts with our method, which aims to learn a socially-aware navigation policy directly from human trajectory data without manually specifying social norms, such as passing on the left.

Crowd simulation methods such as \cite{dutra_gradient-based_2017} and \cite{karamouzas_universal_2014} aim to produce realistic simulations of human movement. However, while the methods use local policies that can be conditioned on a desired destination, they use handcrafted models of human motion with just a few tunable parameters. 

\begin{figure*}[t]
      \centering
      \includegraphics[width=14.5cm]{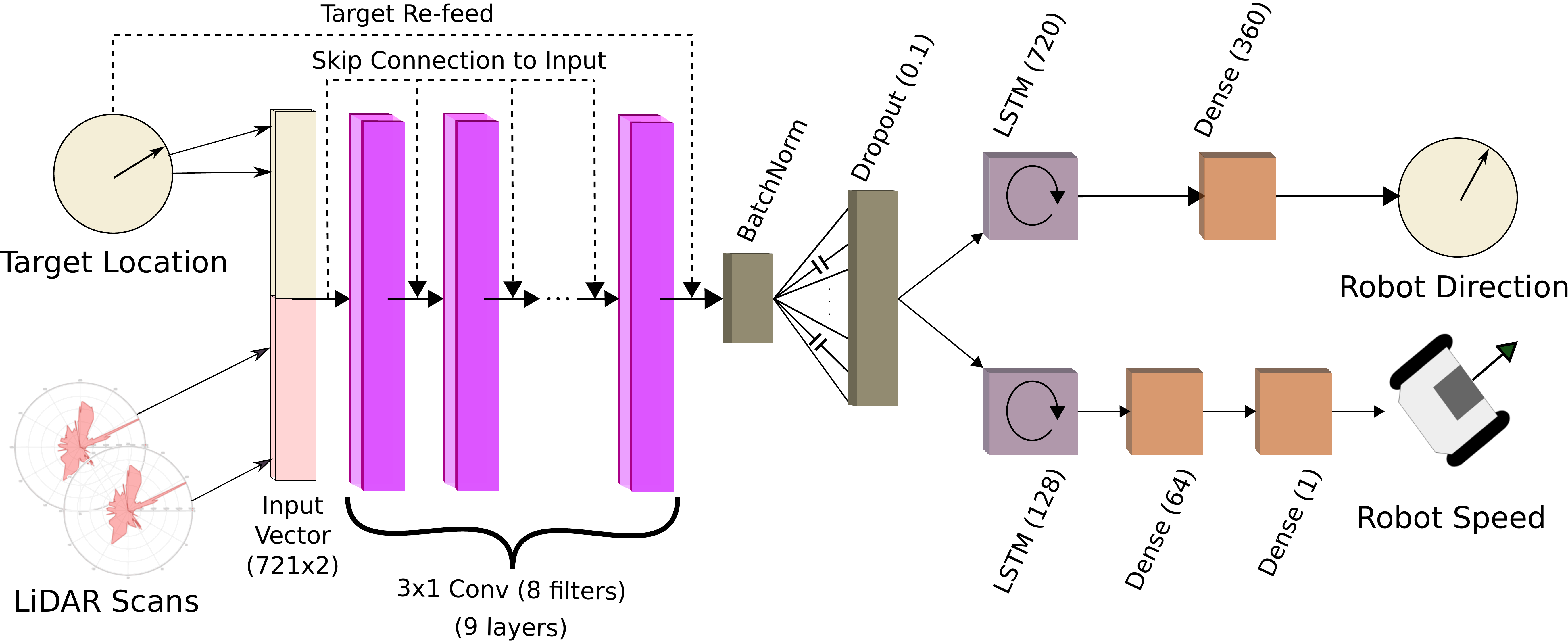}
      \caption{DeepMoTIon network architecture. DeepMoTIon processes the target location and the last two consecutive LiDAR scans via a series of convolutional layers with skip connections followed by batch normalization and dropout. The result is then processed by two separate branches to produce the direction output and the speed output. Each branch consists of LSTM and dense layers.}
      \label{Net_arch}
\end{figure*}

We contrast our problem on human motion imitation with the literature on human motion prediction \cite{alahi2016social,fernando2017soft+}, where the primary focus is predicting a human's future position based on their current location, surroundings, and a history of previous states. The predictions in these methods are not conditioned on a known target, making them unsuitable for the navigation task. With human motion imitation, we assume the robot is given a target location, and its objective is to determine how a human would navigate to it.

In conclusion, methods in the literature show the advantage of imitating humans, although they do not do so from actual human traces. In what follows, we intend to train directly from pedestrian data to alleviate the need for any human modeling through an end-to-end network.

\section{Problem Definition}\label{PD}

Our assumption is that the best way to teach a robot to navigate is to let it learn directly from observing humans' navigation. In this approach, we replace humans one at a time in a pedestrian dataset with our robot equipped with a limited range $360^{\circ}$ LiDAR sensor and let it observe the environment at each time step. Then the robot should learn to mimic the human's navigation for the given observation.

Figure \ref{ALG} shows the different parts of our human motion imitation method. In the offline phase, we train a deep neural network, called DeepMoTIon, based on a pedestrian dataset. In the online phase, the network is given the target direction, target distance, and the last two consecutive LiDAR scans of the environment, and in return the network provides the robot with navigational commands to reach the target while moving similar to humans and ensuring human safety. 

Our model is trained with the ETH pedestrian dataset~\cite{pellegrini2009you} presenting videos of humans navigating in a real-world environment. The dataset contains environment maps and a set~$\chi$ of humans, and for each human $h$ the trajectory~$\zeta_{h}$ that they took through the environment. Each~$\zeta_{h}$ is a sequence of locations~$l_{h,t}$, representing the position of human~$h$ at time~$t$. We use a simulator to estimate the target location~$\tau_{h,t}$, LiDAR scan ~$\mathbf{z}_{h,t}$, and velocity $\mathbf{v}_{h,t}$ at every time step~$t$ for each human replaced by the robot in the dataset, which we then use to train the network to imitate the human trajectories. The simulator uses a manually constructed static obstacle map for each environment along with the annotated human trajectories to approximate the LiDAR scans.

After training, the robot processes its target location and the last two consecutive LiDAR scans via DeepMoTIon to calculate navigational commands that allow it to reach the target safely while moving similarly to the humans in the dataset. This end-to-end learning happens only through observing humans' navigation. The ETH pedestrian dataset is challenging for any autonomous robot due to the dense crowds and sudden changes of pedestrians' directions.

\section{DeepMoTIon}\label{net}

DeepMoTIon is a deep neural network $f(s_{h,t})$ defined as:
\begin{align}\label{deepMotionEq}
    \begin{split}
    &\mathbf{v}_{h,t} = (d_{h,t},v_{h,t}) = f(s_{h,t}), \\
    &s_{h,t} = \begin{bmatrix}
        \mathbf{z}_{h,t-1} & \tau_{h,t}\\
        \mathbf{z}_{h,t} & \tau_{h,t}
    \end{bmatrix},
    \end{split}
\end{align}
\noindent where $s_{h,t}$ is the input state matrix, and $(d_{h,t},v_{h,t})$ is the output action set describing the direction and magnitude of the velocity $\mathbf{v}_{h,t}$. The output direction $d_{h,t}$ is represented as a 360-dimensional vector. This is converted to a scalar heading in degrees by taking the argmax. The input target location $\tau_{h,t}$ is represented in a similar way, being formed by concatenating a 360-dimensional one-hot target direction vector with a scalar target distance to produce a 361-dimensional target location vector. Note that the direction vectors for the input and output can have any number of dimensions, as this simply controls the angular resolution of the direction. We use 360-dimensional vectors because they provide reasonably precise angular resolution and have a convenient translation to degrees.

As shown in Equation (\ref{deepMotionEq}), our network receives the current and previous LiDAR scans in addition to the target location. We found that providing the two LiDAR scans greatly improved the performance of the network compared to only giving the current LiDAR scan. With only one LiDAR scan, the network had difficulty distinguishing between moving and static obstacles.

Our deep neural network architecture is shown in Figure \ref{Net_arch}.
The input to the network is the target location (361$\times$1) and the last two consecutive LiDAR observations (360$\times$2). We concatenate two copies of the target location (361$\times$2) with the $360 \times 2$ LiDAR matrix to create the network's input vector (721$\times$2). The network has 9~convolutional layers, each with 8~($3\times1$) filters and a stride size of 1. The input to each filter was padded to conserve its size, and each filter was followed by a $tanh$ activation function. In this architecture, the skip connections from the input were inspired by classical planning algorithms \cite{groshev2017learning}, such as value iteration and greedy search. However, after the convolutional layers, we re-feed only the raw target location to the network due to its direct correlation to the velocity direction, while the LiDAR scans add minimal value in their raw state. We found through experimentation that only shared convolutional layers were required for the network to correctly deduce the direction and speed from the input state, while adding specialized convolutional layers for each of the two outputs, similar to \cite{groshev2017learning}, reduced its performance.

In addition, for a planning algorithm each state and the corresponding action are tightly related to the previous observations. The LSTM layer was added to the network to
keep some memory of all the previous steps, because these layers have been shown to improve the prediction of future states based on their memory of the past \cite{greff2017lstm}. We later provide a thorough experimental comparison to show the LSTM's necessity. Batch normalization was necessary to assure the boundedness of the input to the LSTM layers. The final dense layers process the LSTM output to provide the direction and speed to be used by the robot.

\subsection{Loss Function}
Our loss function is designed to train the network to output the direction and speed as seen in the human dataset by minimizing the squared error of the speed and the cross entropy error of the output direction.

\begin{figure}[t]
      \centering
      \includegraphics[scale=0.35]{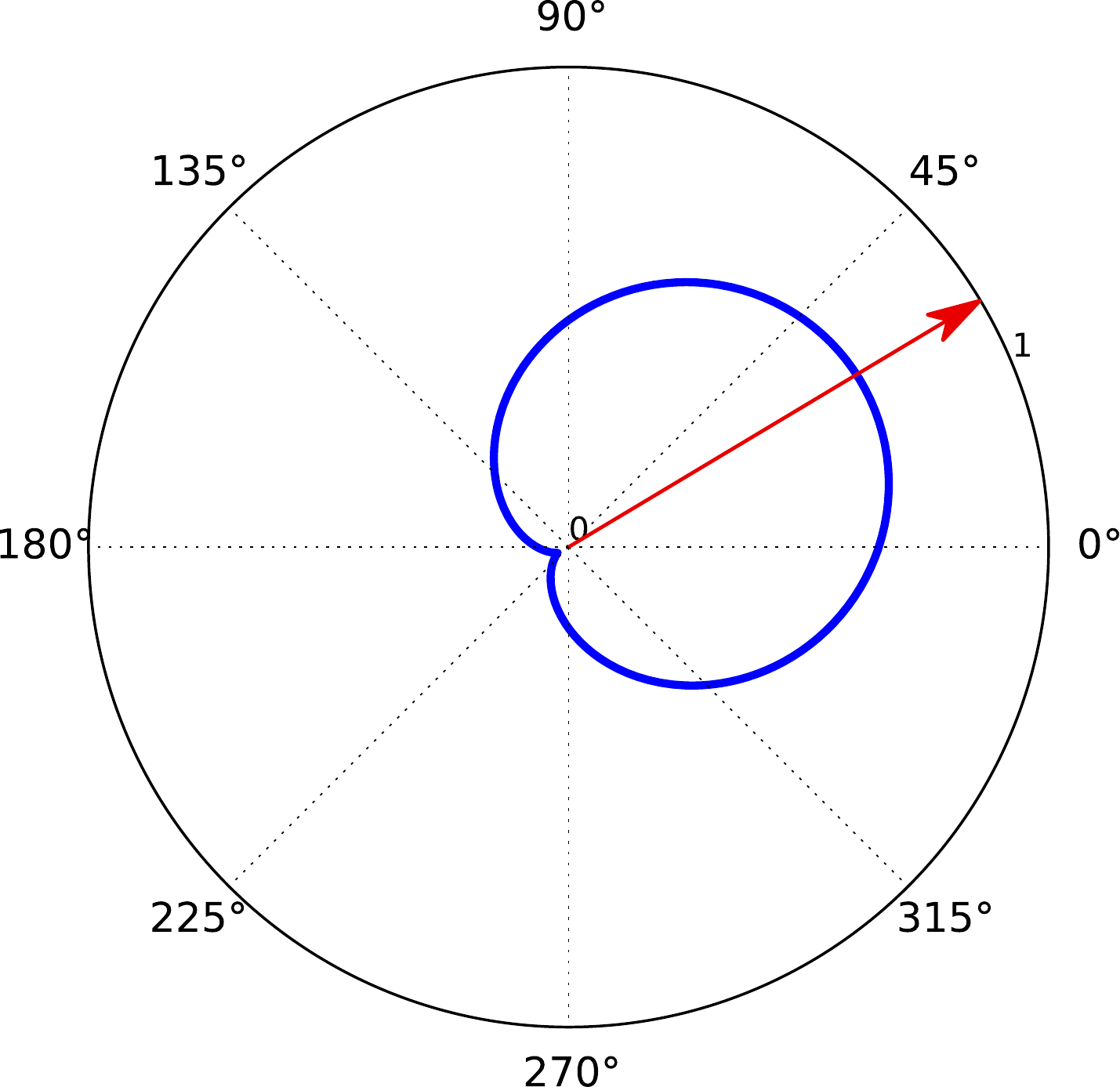}
      \caption{Gaussian distribution (blue) about the human-chosen direction (red) with a standard deviation $\sigma$.}
      \label{GD_d}
   \end{figure}

However, human imitation presents a challenge due to its stochasticity. In fact, two humans might behave differently even with the same observations depending on their personality and other hidden factors. This suggests that the correct direction might be one of many directions in a range about the ground-truth.
As such, it is desirable to penalize the network less for cases where it is close to the ground truth than cases where it is completely wrong. To this end, we model the output direction as a Gaussian distribution about the human-chosen direction with a standard deviation $\sigma$ shown in Figure \ref{GD_d}. The standard deviation $\sigma$ over the human chosen direction is challenging to derive from the pedestrian dataset due to the fact that we only observe a single sample for any given navigational situation. We do not know the range of possible directions that different humans in the same situation might choose because the chosen direction is conditioned on the target position, current location, and observed obstacles, which are different at each timestep for each human. Therefore, we cannot simply compute the standard deviation over all direction choices made by humans in the dataset, and we instead tune $\sigma$ as a model hyperparameter.


As such, our complete loss function for a batch of $N$ training examples can be expressed as follows:

\begin{equation}\label{loss}
    \begin{split}
        &\overbrace{\frac{1}{N}\sum_{i=1}^{N}(v_{i} - \hat v_{i})^2}^\text{Speed Loss} + \overbrace{\frac{1}{N}\sum_{i=1}^{N} H(d_i(\sigma), \hat d_i)}^\text{Direction Loss},
    \end{split}
\end{equation}

\noindent where $v_{i}$ is the actual human speed, $\hat v_{i}$ is the predicted speed, $H(\cdot,\cdot)$ is the cross entropy loss function, $d_i(\sigma)$ is the Gaussian distribution about the human-chosen direction with standard deviation $\sigma$, and $\hat d_i$ is the predicted direction distribution.


\section{Experiments and Results}\label{EX}
To evaluate the performance of DeepMoTIon when imitating humans, we conducted experiments on the ETH BIWI walking pedestrians dataset \cite{pellegrini2009you}. The dataset provides annotated trajectories of 650 humans recorded over 25 minutes of time on two maps. We randomly assigned 2/3 of the data to the training set and 1/3 to the test set. To avoid overfitting and allow the network to generalize to unseen maps, the training data was augmented by replicating each path while rotating the map at random angles.

The dense crowds and sudden changes of pedestrian direction in this dataset make it sufficiently challenging for our experiments. We conducted two types of experiments on this dataset. First, we assessed the different components of the network, and then we compared it with other benchmark methods in terms of human imitation, safety, and target reachability. 

It should be noted that our algorithm runs in real-time despite the depth of the network due to the low dimensionality of its input state vector. GPU was only used for training. During testing, the algorithm runs in real-time on a single core of a 3.3 GHz CPU (0.084 seconds per forward pass), which can be set up easily on a mobile ground robot.

\subsection{Benchmarks\label{bench}}
To assess the performance of DeepMoTIon, we ran several baseline algorithms in our experiments for comparison. These include two deep learning algorithms as well as a human-aware navigation method from the literature.
\begin{enumerate}
\item {\bf Generalized Reactive Planner (GRP)}: GRP is a deep neural network architecture composed of multiple convolutional layers, each with a skip connection from the input, followed by fully connected layers to the output \cite{groshev2017learning}. In addition, the input of GRP is a concatenation of the observation and target. GRP is trained to learn reactive policies that allow the robot to imitate a planning algorithm.
\item {\bf End-to-end Motion Planning (EMP)}: EMP is another deep neural architecture that relies on a relatively small number of deep convolutional layers and two residual shortcut connections \cite{pfeiffer2017perception}. Moreover, the target in this network is provided after the convolutional layers. EMP is trained to learn a navigation algorithm.
\item {\bf Social Force Model (SFM)}: SFM calculates a set of imaginary `social' forces that govern the human motion in a crowd \cite{ferrer2017robot}. These forces can be grouped as repulsive to obstacles and other humans as well as attractive to the target.
\end{enumerate}

The optimizer for all deep architectures in our experiments (GRP, EMP, and DeepMoTIon) was Adadelta, with a learning rate of 1 and an L2 regularization weight of 0.001.
 
\subsection{Metrics}
To assess our network and the benchmark algorithms, we compared their performance when trying to navigate from a start position to the final target. The start and final positions are chosen from the dataset, where the simulator replaces one of the humans with a robot and compares the resemblance of their paths, as well as the safety of the robot and other humans in the environment. Formally, we compare the performance of our network and the benchmark algorithms based on the following metrics:

\begin{enumerate}
\item \textbf{Squared Path Difference (SPD)}: The trajectories of the robot and the corresponding human are modeled as discrete-time trajectories $T_{r,0..n}$ and $T_{h,0..m}$ respectively. The squared path difference can then be expressed as
\begin{equation}
\sum^{\max(n,m)}_{i=1}{||T_{r,i}-T_{h,i}||^2},
\end{equation}
\noindent where the last location in the shorter path is compared with the remaining steps of the other. This metric indirectly penalizes the difference in length between the robot's and the human's paths.

\item \textbf{Dynamic Time Warping (DTW):} \textit{DTW} is a metric described in ~\cite{10011461178} to measure the similarity between two temporal sequences, which may vary in speed. The metric finds the optimal time warp to match the segments of the two paths and measures the similarity between them following that warp. While \textit{SPD} reflects an algorithm's ability to replicate both direction and speed, \textit{DTW} compares the two paths irrespective of their speed. For instance, similarities in walking could be detected using \textit{DTW}, even if the imitating robot was walking faster or slower than the human, or if there were accelerations and decelerations over the course of the navigation. 

\item \textbf{Proximity:} \textit{Proximity} is the closest distance the robot comes to a human on its path. In the case of any collision along the path, it is assigned a value of $0$. We report the average \textit{proximity} over all the test cases. 

\item \textbf{Number of Collisions:} The number of times the robot collides with a human while navigating in the environment.

\item \textbf{Target:} The percentage of trials where the robot reaches the goal within the 400-step threshold.
\end{enumerate}

It should be noted that unlike most human-aware navigation papers \cite{ferrer2017robot,vasquez2013human}, we are reporting the average number of collisions as a comparison metric. However, when implemented in real-world settings, a low-level obstacle avoidance controller is to be added to the algorithm to assure complete human safety and accommodate for any failures similar to~\cite{kim2016socially}.

 \begin{figure*}[t]
      \centering
      \vspace{0.2cm}
      \includegraphics[clip, trim={0cm 0cm 0cm 0cm},scale=0.80]{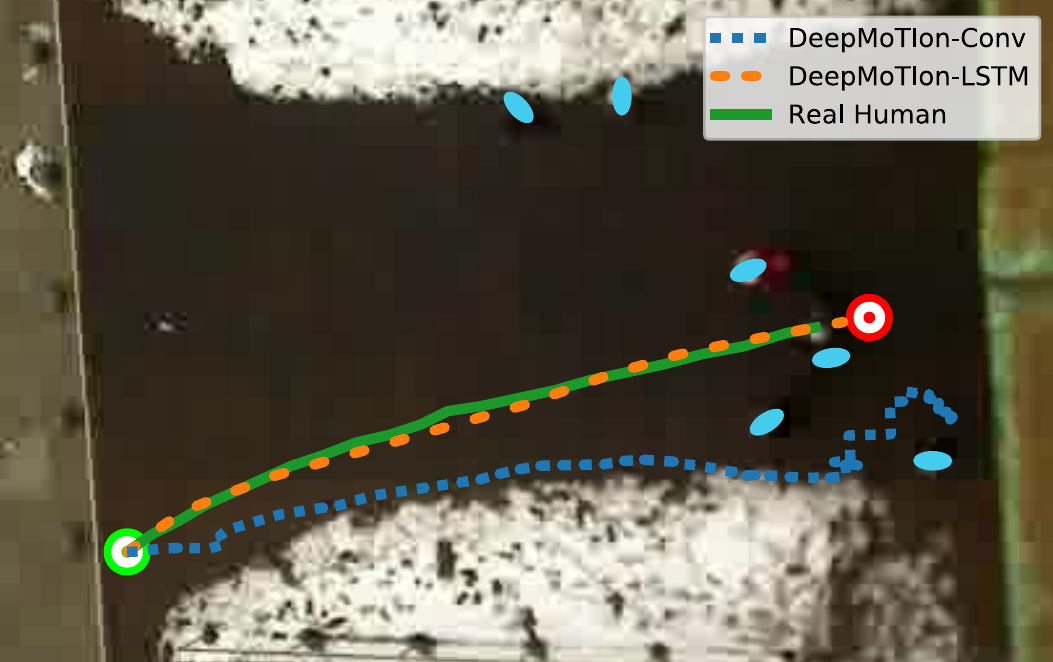}\hspace{1em}%
      \includegraphics[clip, trim={0cm 0cm 0cm 0cm},scale=0.80]{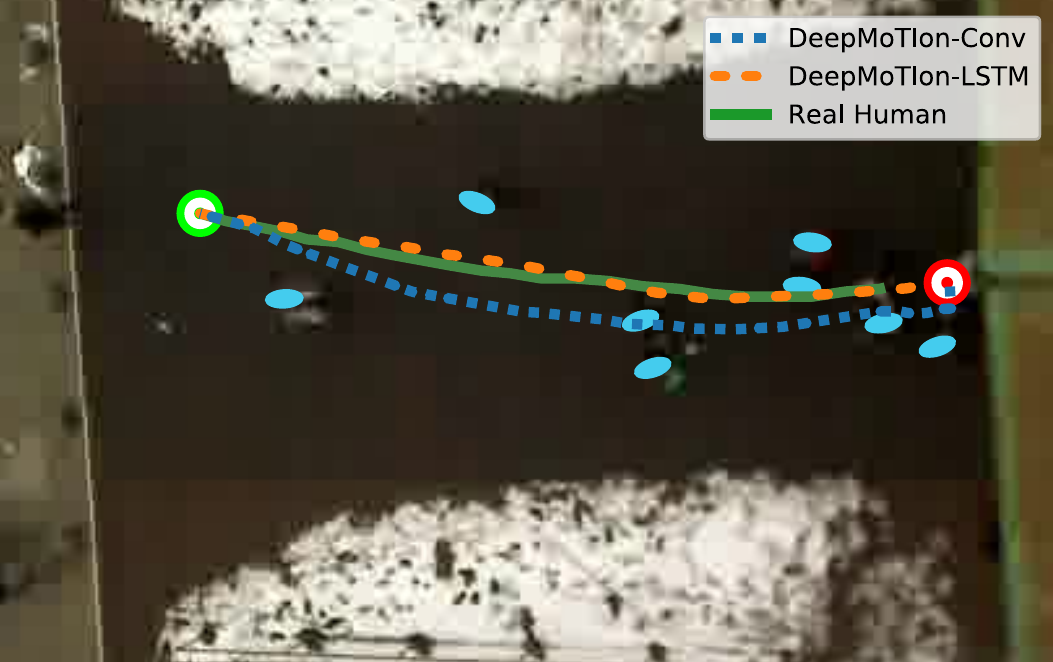}
          \caption{Demonstrations of our network navigating from the start location (green) to the goal location (red). The left figure shows a scenario where $\text{DeepMoTIon}_{Conv}$ fails, while $\text{DeepMoTIon}_{LSTM}$ finds the target. The right figure shows a case where both networks  reach  the target, with the LSTM variant imitating the human more closely.}
      \label{demo}
   \end{figure*}


\subsection{DeepMoTIon Variants}

To study the necessity of the LSTM layer, we tested two variants of the network, where we refer to $\text{DeepMoTIon}_{LSTM}$ as the network with the architecture explained before and $\text{DeepMoTIon}_{Conv}$ as the network without any LSTM layers. More convolutional layers were added to $\text{DeepMoTIon}_{Conv}$ to accommodate for the depth difference. Both variants of $\text{DeepMoTIon}$ were trained with the full loss function shown in Equation (\ref{loss}) and with a fixed $\sigma=5$.

Table \ref{comp} shows that $\text{DeepMoTIon}_{LSTM}$  imitates humans better than $\text{DeepMoTIon}_{Conv}$ and exhibits a better performance in all metrics. $\text{DeepMoTIon}_{LSTM}$ outperforms $\text{DeepMoTIon}_{Conv}$ with regard to path difference (\textit{SPD} and \textit{DTW}) and safety (\textit{proximity} and the number of collisions) and reaches the target in $100\%$ of the trials. The results prove the necessity of the LSTM layer and the fact that having two consecutive LiDAR scans fed into the network does not work on its own.

Figure \ref{demo} also shows example robots navigating with the two variants of DeepMoTIon's architecture.
Figure \ref{demo} (left) shows an example where $\text{DeepMoTIon}_{conv}$ is accumulating error throughout its path and finally misses the target. This behavior was observed throughout the trials on many occasions, which explains the difference in performance of the two networks. The figure also shows that $\text{DeepMoTIon}_{LSTM}$ was able to follow the human path all the way to the target. These observations suggest the necessity of the LSTM layer for the network to acknowledge the existence of the error and correct it when required.
Figure \ref{demo} (right) shows an example where both networks reached the target, with the LSTM variant imitating the human more closely.


\begin{table}[t]
\vspace{0.2cm}
\centering
\caption{Performance Metrics Comparison}
\label{comp}
\begingroup
\setlength{\tabcolsep}{4pt} 
\resizebox{\columnwidth}{!}{%
\begin{tabular}{l|cccccc}

\cmidrule[\heavyrulewidth]{2-6}

                                & \textit{SPD} & \textit{DTW}& \textit{Proximity} & \textit{Collisions} & \textit{Target} \\ 
                      \midrule
 
$\textbf{DeepMoTIon}_{LSTM}$              & 151      & 39          & 0.31               & 0.67                  & 100\%            \\ 
$\textbf{DeepMoTIon}_{Conv}$              & 732           & 131          & 0.25               & 0.89                & 69\%            \\ 
\text{SFM  \cite{ferrer2017robot}       } & 3817             & 51           & 0.29               & 0.26                & 100\%            \\ 
\text{EMP  \cite{pfeiffer2017perception}} & 15437           & 1187         & 0.001               & 7.69                & 32\%            \\ 
\text{GRP  \cite{groshev2017learning}   } & 334             & 52          & 0.18               & 0.78                & 84\%
\end{tabular}
}
\endgroup
\end{table}

\subsection{Comparison with Benchmarks}
Table \ref{comp} shows the performance of each algorithm we evaluated on the test set. We randomly assigned 2/3 of the data to the training set and 1/3 to the test set. The numbers in the table are averages over all test examples. The results illustrate the ability of $\text{DeepMoTIon}_{LSTM}$ to imitate humans better than the other benchmark algorithms. \textit{SPD} and \textit{DTW} show that $\text{DeepMoTIon}_{LSTM}$ has the lowest path difference among all the tested algorithms, with the next best algorithm, GRP, showing more than double the path difference. 

We note that $\text{DeepMoTIon}_{LSTM}$ and SFM both reach the target on 100\% of the trials, while other networks often fail. EMP reaches the target in 32\% of trials and GRP reaching the target in 84\% of the trials.
The \textit{proximity} parameter shows $\text{DeepMoTIon}_{LSTM}$ keeps an average \textit{proximity} of $0.31m$ to any human, which is safer than the other benchmark algorithms. SFM keeps an average \textit{proximity} of $0.29\text{m}$ despite explicitly weighting its repulsive force to humans higher than the other social forces. 

With regard to the number of collisions, SFM has the lowest rate among all the algorithms. This can be explained by the ability of the algorithm to stop in the case of dense crowds, while all the other networks were not trained on any human demonstration that exhibited that type of behavior. We expect DeepMoTIon to learn to stop and avoid collisions better when trained on more pedestrian data showing a wider set of possible navigation scenarios. We also expect that with enough training data in a variety of environments with different features, e.g., crowd density or crowd speed, the network can learn to perform well even when placed in a new environment that it had not been trained on.

Finally, we note that our network was able to navigate even with a LiDAR range other than the one it was trained on. All the algorithms above were trained and tested on a LiDAR with a $30m$ range. To show the ability of our network to generalize to different ranges, we tested its performance with a $6m$ LiDAR range without any retraining. The network was still able to reach the target on 97\% of the trials, with an increase in \textit{DTW} to $47$ and a decrease in the number of collisions to $0.51$. The decrease in collisions was expected, as the network is observing obstacles in locations that were supposed to be free, and thus the robot becomes more careful.


\section{Conclusion and Future Work}\label{End}

We introduced a novel deep imitation learning framework and studied its performance when learning to navigate from human traces. We trained the deep network to predict robot command velocities from raw LiDAR scans without the requirement of any preprocessing or classification of the surrounding objects. 
Our experiments showed DeepMoTIon's ability to generate navigation commands similar to humans and plan a path to the target on all test sets, outperforming all the benchmarks on path difference (\textit{SPD} and \textit{DTW}) and \textit{proximity} metrics, and all except SFM on the number  of collisions. In addition, we presented a novel loss function to train the network. The loss function allowed us to accommodate for human motion stochasticity while at the same time enabling the robot to navigate safely. Finally, we presented a comparative assessment that showed the necessity of an LSTM layer for a planning algorithm via a deep neural network, where the robot navigating with the non-LSTM variant of our network was led astray on many test cases. 

In the future, we plan to train the network to navigate using raw images instead of LiDAR scans, where we believe the larger bandwidth of the data can help the network understand human motion from their point of view. However, unlike DeepMoTIon, special consideration has to be taken when training the network with images to provide a navigation model that runs in real time when implemented on a mobile platform. 

\bibliographystyle{abbrv}
\balance
\bibliography{main}

\end{document}